\documentclass[letterpaper]{article} 
\usepackage{aaai25}  
\usepackage{times}  
\usepackage{helvet}  
\usepackage{courier}  
\usepackage[hyphens]{url}  
\usepackage{graphicx} 
\usepackage{natbib}  
\usepackage{caption} 
\usepackage{algorithm}
\usepackage{algorithmic}
\usepackage{amsmath,amssymb}
\usepackage{tabularx}
\usepackage{multirow}
\usepackage{seqsplit}
\newsavebox\CBox
\def\textBF#1{\sbox\CBox{#1}\resizebox{\wd\CBox}{\ht\CBox}{\textbf{#1}}}



\usepackage{newfloat}
\usepackage{listings}
\DeclareCaptionStyle{ruled}{labelfont=normalfont,labelsep=colon,strut=off} 
\floatstyle{ruled}
\newfloat{listing}{tb}{lst}{}
\floatname{listing}{Listing}
\title{EventMamba: Enhancing Spatio-Temporal Locality with State Space Models for Event-Based Video Reconstruction}
\author{
    Chengjie Ge,
    Xueyang Fu,
    Peng He,
    Kunyu Wang,
    Chengzhi Cao,
    Zheng-Jun Zha\thanks{Corresponding author.}
}
\affiliations{
    University of Science and Technology of China, China\\

    cjge@mail.ustc.edu.cn,
    xyfu@ustc.edu.cn,
    \{hp0618,kunyuwang,chengzhicao\}@mail.ustc.edu.cn,
    zhazj@ustc.edu.cn
}

\usepackage{bibentry}

\begin{document}

\maketitle

\begin{abstract}
Leveraging its robust linear global modeling capability, Mamba has notably excelled in computer vision. Despite its success, existing Mamba-based vision models have overlooked the nuances of event-driven tasks, especially in video reconstruction. Event-based video reconstruction (EBVR) demands spatial translation invariance and close attention to local event relationships in the spatio-temporal domain. Unfortunately, conventional Mamba algorithms apply static window partitions and standard reshape scanning methods, leading to significant losses in local connectivity. To overcome these limitations, we introduce EventMamba—a specialized model designed for EBVR tasks. EventMamba innovates by incorporating random window offset (RWO) in the spatial domain, moving away from the restrictive fixed partitioning. Additionally, it features a new consistent traversal serialization approach in the spatio-temporal domain, which maintains the proximity of adjacent events both spatially and temporally. These enhancements enable EventMamba to retain Mamba’s robust modeling capabilities while significantly preserving the spatio-temporal locality of event data. Comprehensive testing on multiple datasets shows that EventMamba markedly enhances video reconstruction, drastically improving computation speed while delivering superior visual quality compared to Transformer-based methods.
\end{abstract}

%

\section{Introduction}
Event cameras, also known as neuromorphic cameras, draw inspiration from biological systems and offer substantial advancements over traditional visual sensors. They provide exceptional temporal resolution (1 $\mu$s), superior dynamic range (140 dB), and ultra-low power consumption (5 mW)~\cite{gallego2020event,delbruck2010activity,benosman2013event,fu2024event}. Unlike conventional cameras, event cameras capture data asynchronously and sparsely, which complicates direct interpretation and integration with standard computer vision techniques. To address this, converting event data into more conventional intensity images is essential for bridging the technological divide in computer vision applications.

Over the past decades, deep learning has achieved remarkable progress in computer vision~\cite{ge2022learning,ge2024neuromorphic,zhang2023fgnet,zhang2024genseg,zhang2023baproto,xiao2022stochastic,li2023edge,shi2022athom,peng2024lightweight,wang2024towards}, especially in event-based video reconstruction (EBVR)~\cite{zhu2022event,rebecq2019events,cadena2023sparse,scheerlinck2020fast,gallego2020event}. Current methods typically utilize Convolutional Neural Networks (CNNs) or Transformers to reconstruct frames from event data. While CNNs focus on local details, often at the expense of global context, this can lead to increased susceptibility to noise and blurring, resulting in unclear visual outputs~\cite{jang2021noise}. Conversely, Transformers excel in capturing extensive non-local information through their self-attention mechanisms~\cite{weng2021event,xu2024demosaicformer}. However, this approach scales quadratically with the input size, noted as $O(n^{2})$. This scaling issue is especially challenging for high-resolution data, like the 1280$\times$720 output from Prophesee EVK4 cameras, due to its high computational demands and the difficulty of deploying on devices with limited resources.

\begin{figure*}
\centering
\includegraphics[width=\linewidth]{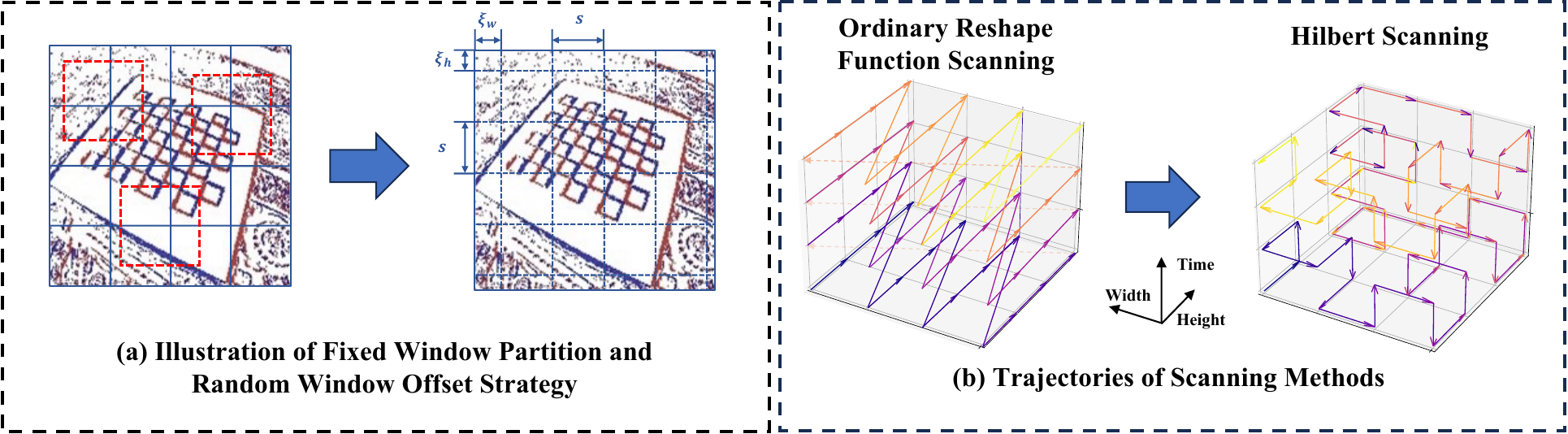}
\caption{(a) Example illustrating the loss of locality in the fixed window strategy (spatial locality loss in the red box), and our proposed Random Window Offset solution.
(b) Demonstration of the loss of spatio-temporal locality in conventional space-filling curves contrasted with our introduced Hilbert space-filling curve technique.}
\label{pic:1}
\end{figure*}

Recently, the Mamba module, particularly within the State Space Model (SSM) framework, has introduced a groundbreaking approach to address previous challenges~\cite{gu2021efficiently}. As a subset of SSMs, the advanced Mamba modules~\cite{gu2023mamba} have shown substantial advancements by employing a sophisticated selection mechanism and hardware optimizations~\cite{zhu2024vision,liu2024vmamba,xing2024segmamba,liu2024swin}. However, these vision Mamba modules are not directly suitable for the EBVR task due to two main reasons. Firstly, EBVR tasks require translation invariance in the spatial domain, necessitating a location-independent approach to map events to video frames. Traditional vision Mamba networks, such as VisionMamba and VmambaIR, use fixed non-overlapping windows that constrain SSM operations to these local windows, thereby imposing spatial priors inappropriate for EBVR tasks, leading to a loss of translation invariance and incomplete local relationship capture. Secondly, event data in the spatio-temporal domain is critical, and the common practice of flattening temporal features into a one-dimensional sequence processed recursively disrupts the natural spatio-temporal event relationships, resulting in a loss of local information.

To overcome these challenges, we introduce EventMamba, a specialized SSM network designed for high-speed EBVR tasks. In the spatial domain, EventMamba employs a novel random window offset (RWO) strategy, which uses randomly offset windows to encompass the entire feature map, rather than restricting it to fixed partitions (see Figure~\ref{pic:1}(a)). This RWO strategy ensures preservation of translation invariance and more comprehensive local relationship mapping. In the spatio-temporal domain, EventMamba employs a unique Hilbert Space Filling Curve (HSFC) scanning mechanism. Compared to other space-filling curves, the Hilbert curve exhibits superior locality preserving properties and a lower space-to-linear ratio~\cite{chen2022efficient,bauman2006dilation,wu2024rainmamba}. This implies that the Hilbert curve is more effective at retaining the local characteristics of the original data and offers higher efficiency when mapping multi-dimensional space to a one-dimensional linear sequence. EventMamba leverages these inherent advantages of Hilbert and trans-Hilbert curves to convert spatio-temporal pixels into a one-dimensional sequence along the curve trajectory, enabling a fine-grained and locality-preserving recovery of event data (see Figure~\ref{pic:1}(b)).  Through these innovations, EventMamba maintains the spatial and temporal locality of event data while leveraging the powerful linear global modeling capabilities of Mamba.

In summary, our contributions are as follows:
\begin{itemize}
\item We conduct a critical analysis of the limitations of existing Transformer and CNNs-based methods and introduce EventMamba, a pioneering model that integrates State Space Models for event-based video reconstruction.
\item We elaborately design a random window offset strategy for reconstruction tasks to compensate for the loss of translation invariance caused by previous vision Mamba models when using fixed partitioned windows, thereby better modeling local information in the spatial domain.
\item We design a Hilbert Space Filling Curve mechanism tailored for EBVR tasks to address the disruption of spatio-temporal locality in previous vision Mamba models, significantly enhancing the model's ability to capture spatio-temporal relationships.
\end{itemize}

Our comprehensive experimental results show that EventMamba markedly outperforms existing models, enhancing both subjective and objective performance metrics. On the IJRR dataset~\cite{mueggler2017event}, EventMamba increases the SSIM value by 2.9\% compared to previous state-of-the-art approaches.

\section{Related Works}

\subsection{Event-based Video Reconstruction}
Early studies on EVBR tasks were primarily based on physical priors of the event stream, which were significantly limited by specific conditions of the photographic scenes~\cite{kim2008simultaneous,lagorce2016hots,munda2018real,chen2020event,gehrig2021dsec,schaefer2022aegnn,shiba2022secrets,tulyakov2022time,freeman2023asynchronous}. Kim~\textit{et al.} developed a method based on the Kalman filter to reconstruct gradient video frames from a rotating event camera, and used Poisson integration techniques to recover luminance frames with temporal dimensions~\cite{kim2008simultaneous}. Bardow~\textit{et al.} proposed a variational energy minimization framework that allows simultaneous recovery of video frames and dense optical flow from the sliding window of an event camera~\cite{barua2016direct}.  Zhang~\textit{et al.} formulated the event-based video frame reconstruction task as an optical flow-based linear inverse problem, demonstrating that this approach could generate luminance video frames of quality comparable to those trained with deep neural networks, without the need for training deep networks~\cite{zhang2021formulating}.

\begin{figure*}
\centering
\includegraphics[width=0.98\linewidth]{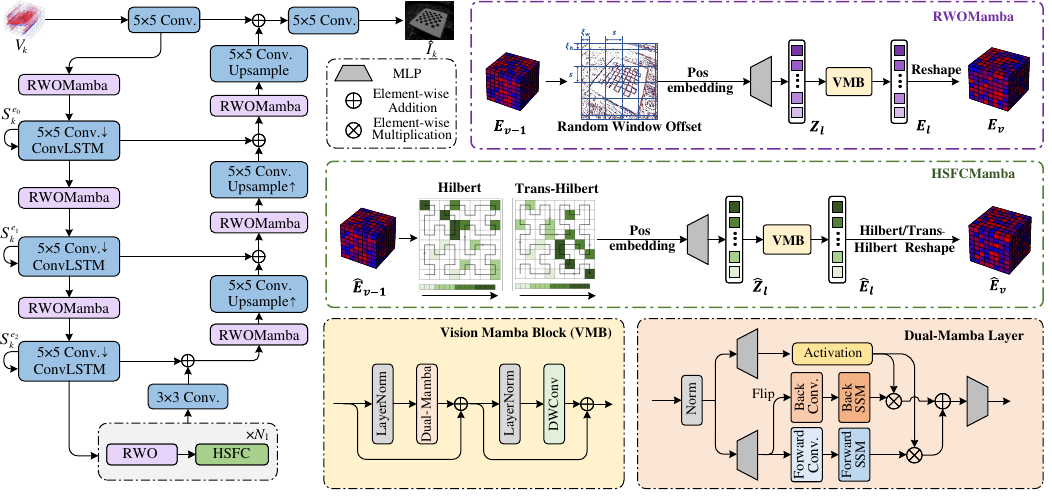}
\caption{The EventMamba architecture is U-Net-like, processing event voxels ($V_k$) to predict intensity images. It incorporates two key components: RWOMamba and HSFCMamba, which are designed to maintain the translation invariance and spatio-temporal locality of event features, respectively. The number of $N_1$ is set to 2 in our EventMamba architecture.}
\label{pic:2}
\end{figure*}

In recent years, with the development of deep learning, data-driven neural network algorithms have made significant breakthroughs in the field of EBVR. Rebecq~\textit{et al.} developed E2VID, a model for event-based video frame reconstruction that combines the advantages of CNNs and Recurrent Neural Network (RNNs), improving the quality of video frame reconstruction through controlled incremental updates of event sequences~\cite{rebecq2019events}. To address the vanishing gradient problem in long sequence data processing, Rebecq~\textit{et al.} later introduced E2VID+, which incorporates multi-layer ConvLSTM units to stabilize gradients during backpropagation~\cite{stoffregen2020reducing}. Cadena~\textit{et al.} used spatially-adaptive denormalization (SPADE) layers in the E2VID framework and proposed that the SPADE module enhanced the quality of reconstructed frames in the video~\cite{scheerlinck2020fast}. Wen~\textit{et al.} were among the first to apply the Transformer architecture to the field of event-based video reconstruction, combining the local feature extraction capabilities of CNNs with the global information processing advantages of Transformer to further enhance the quality of event-based video reconstruction~\cite{weng2021event}. Zhu~\textit{et al.} proposed a novel EBVR network based on spiking neural networks, their meet the comparable performance with ANN methods while saves the energy consumption~\cite{zhu2022event}. Cadena~\textit{et al.} considered the sparsity of the event stream in event-based video reconstruction and were the first to employ stacked sparse convolutional modules in the reconstruction network, effectively reducing the network's complexity~\cite{cadena2023sparse}. 

\subsection{State Space Model}

State Space Models (SSMs), first introduced in the S4 model~\cite{gu2021efficiently}, model global information more efficiently than CNNs or Transformers. S5~\cite{smith2022simplified} reduced the complexity to linear levels using MIMO SSMs and parallel scanning. H3~\cite{mehta2206long} added gating units to enhance expressiveness, enabling it to compete with Transformers in language modeling. SSM-ViT~\cite{zubic2024state} significantly enhances performance and training speed at high frequencies by incorporating the SSMs into event-based vision. Mamba~\cite{gu2023mamba} introduced an input-adaptive mechanism, outperforming similarly-scaled Transformers in inference speed, throughput, and overall performance. 

Motivated by the success of Mamba in language modeling, various Mamba-based models have been proposed for vision tasks~\cite{zhu2024vision, liu2024vmamba, xing2024segmamba, shi2024vmambair,li2024videomamba,wu2024rainmamba}. However, these models often directly apply fixed partitioned windows to vision tasks without fully considering the unique characteristics of event-based tasks. EBVR tasks require higher translation invariance and spatio-temporal locality compared to classification and segmentation tasks, as previously discussed. This is because EBVR tasks necessitate feature integration across both temporal and spatial dimensions for pixel-level regression. To effectively address EBVR tasks, it is crucial to develop a mechanism that overcomes the limitations of current vision Mamba models in terms of translation invariance and spatio-temporal locality.

\section{Methodology}
In this section, we introduce the fundamental concepts of SSMs. We then provide a detailed explanation of how we integrate SSMs with the EBVR task. This includes a description of the model framework, modular design, training strategies, and the loss functions employed.

\subsection{Preliminaries}
State Space Sequence Models and the Mamba framework are based on linear dynamics principles, transforming a one-dimensional sequence $x(t) \in \mathbf{R}$ through a hidden state space $h(t) \in \mathbf{R}^N$ to produce an output $y(t)$. The transformation involves matrices $\mathbf{A} \in \mathbf{R}^{N \times N}$ (state evolution), $\mathbf{B} \in \mathbf{R}^{N \times 1}$ (input-to-state), and $\mathbf{C} \in \mathbf{R}^{1 \times N}$ (state-to-output). The system dynamics are described by:
\begin{align}
&h'(t)= \mathbf{A}h(t) + \mathbf{B}x(t),\\
&y(t)= \mathbf{C}h'(t).
\end{align}

The S4 and Mamba modules adapt these models to discrete time, using a time scale $\Delta$ to convert $\mathbf{A}$ and $\mathbf{B}$ to their discrete counterparts $\bar{\mathbf{A}}$ and $\bar{\mathbf{B}}$, employing the Zero-Order Hold (ZOH) method~\cite{karafyllis2011nonlinear} :
\begin{align}
\bar{\mathbf{A}} &= \exp(\mathbf{\Delta A}),\\
\bar{\mathbf{B}} &= (\mathbf{\Delta A})^{-1}(\exp(\mathbf{\Delta A} - \mathbf{I})) \cdot \mathbf{\Delta B}.
\end{align}
Finally, the output is obtained through global convolution:
\begin{align}
h_t &= \bar{\mathbf{A}}h_{t-1} + \bar{\mathbf{B}}x_t,\\
y_t &= \mathbf{C}h_t.
\end{align}
The output is generated by a structured convolution:
\begin{align}
\bar{\mathbf{K}} &= (\mathbf{C}\bar{\mathbf{B}}, \mathbf{C}\bar{\mathbf{A}}\bar{\mathbf{B}}, \ldots, \mathbf{C}\bar{\mathbf{A}}^{\mathbf{M-1}}\bar{\mathbf{B}}),\\
\mathbf{y} &= x * \bar{\mathbf{K}},
\end{align}
wherein $\mathbf{M}$ represents the length of the sequence $x$, and $\bar{\mathbf{K}}\in \mathbf{R}^M$ denotes a structured convolutional kernel.

\subsection{Event Representation}
We consider an event stream ${e_i}$ containing $N_E$ events over a duration of $T$ seconds, where each event $e_i = (x_i, y_i, t_i, p_i)$ encodes the position $(x_i, y_i)$, timestamp $t_i$, and polarity $p_i$ of the $i$-th brightness change detected by the sensor. The goal is to generate a stream of video frames ${\hat{I}_k}$ from the same $T$-second interval, where each video frame $\hat{I}_k \in [0, 1]^{W \times H}$ represents a 2D grayscale representation of the scene's absolute brightness. $H$ and $W$ represent the height and width. The proposed method limits each generated video frame to rely solely on past events.

We sort events into groups corresponding to the timestamps of the video frames. Given $N_I$ frames, each identified by a timestamp $s_k$, we define the $k$-th event group as:
\begin{equation}
G_k \dot{=} \{e_i \mid s_{k-1} \leq t_i < s_k\}, \quad \text{for} \ k = 1, \ldots, N_I.
\end{equation}
To input these organized events into CNNs, the events are amassed into a voxel grid $V_k \in \mathbf{R}^{W \times H \times B}$, where $W$ and $H$ correspond to the grid's dimensions, and $B$ is the number of distinct bins. The timestamp $t_i$ of each event is normalized to the range $[0, B-1]$, yielding the normalized timestamp $t^*_i$:
\begin{equation}
t^*_i = \frac{(B-1)(t_i - T_k)}{\Delta T}.
\end{equation}
We then use an interpolation method to distribute the polarity contribution of each event across the nearest two voxels in the temporal dimension:
\begin{equation}
V_k(x, y, t) = \sum p_i \max(0, 1 - |t - t^*_i|) \delta(x - x_i, y - y_i).
\end{equation}
In the experiments, $B = 5$ is selected as the count of discrete intervals for the time dimension, resulting in each event cluster being represented by a voxel grid with dimensions $W \times H \times 5$, which can then be fed into DNNs for further processing and analysis.

\subsection{Network Structures}
\textbf{Random Window Offset Mamba.}
In previous iterations of the vision Mamba model, image features were divided into non-overlapping windows and processed through the Mamba module for experimental purposes. However, for reconstruction tasks, all window partitions contain equally important information. Therefore, the use of fixed window partitions results in a loss of translation invariance. To address this issue, we propose the Random Window Offset (RWO) strategy to endow the EventMamba model with translation invariance and to fully utilize the local relationships in the spatial domain as shown in Figure~\ref{pic:2}. The vision mamba block (VMB) is a residual structure that combines a dual-mamba layer, inspired by the VisionMamba~\cite{zhu2024vision}, with a depth-wise convolution (DWConv)~\cite{chollet2017xception} layer. The computation is defined as follows:
\begin{align}
\begin{split}
&z_l=\text{MLP}(\text{RWO}(E_{v-1};s,\xi_h^l,\xi_w^l)), \; (\xi_h^l,\xi
_w^l)\sim\mathbb{U}(\mathfrak{R}_s),\\
&E_l=\text{VMB}(z_l),\\
\end{split}
\end{align}
where $E_{v-1}$ represents the input event feature from the previous stage, $s$ is the spatial size of local window, $z_l$ denotes the output sequence after the MLP layer, and $E_l$ denotes the output sequence after the VMB layer. $\mathfrak{R}_s$ includes all possible offsets within the uniform distribution $\mathbb{U}(\mathfrak{R}_s)$. The representation of $\mathfrak{R}_s$ can be simplified as:
\begin{equation}
\mathfrak{R}_s:=[0,\dots,s-1]\times[0,\dots,s-1].
\end{equation}
In the training process, the symbol $\times$ denotes the Cartesian product. The parameters $(\xi_h^l,\xi
_w^l)$ are considered as independently and identically distributed random variables sampled from the uniform distribution $\mathbb{U}(\mathfrak{R}_s)$. Assuming the total number of Mamba layers is $N$, the random displacements $\{(\xi_h^l,\xi
_w^l)\}_{l=0}^{N-1}$ are intentionally designed to be independent to ensure the maintenance of faithful locality and translation invariance at the layer level. Through this approach, although each individual layer considers all possible displacements, only one set of sampled displacements $\{(\xi_h^l,\xi
_w^l)\}_{l=0}^{N-1}$ is required for each forward propagation, thus the training time remains consistent with that of fixed window partitioning.
Following the RWO strategy, the event features are input into the VMB module and reshape to its original shape $E_v$. 

Regarding the testing process, the conventional strategy is to use layer-wise expectation inspired by Dropout~\cite{srivastava2014dropout} to approximate the overall output of the model, which is expressed as:
\begin{equation}
EM^{test}(x)=\underset{s}{\mathbb{E}}[EM(x;s,\xi_h^l,\xi_w^l)],
\end{equation}
where $EM$ stands for our EventMamba network for simplicity.
However, for our task, using layer-wise expectation requires traversing all combinations of random windows, this computation method will greatly increase the computational burden, which contradicts our original intention of introducing the Mamba model. Therefore, we use another Monte Carlo approximation to estimate the output of the model, which is expressed as:
\begin{equation}
EM^{test}(x)\approx\frac{1}{M}\sum_{i=1}^{M}[EM(x;s_i,\xi_h^l,\xi_w^l)].
\end{equation}
It may seem that the testing time would scale linearly with $M$, the number of averaged forward passes. However, modern accelerators can perform multiple forward passes concurrently, significantly reducing the testing time. This acceleration is achieved by transferring an input to the GPU(s) and creating a mini-batch that consists of the same input repeated multiple times. EventMamba performs RWO operations independently along the batch dimension, allowing for parallel processing. After a single forward pass through EventMamba, the Monte Carlo estimate is obtained by averaging over the mini-batch. For readability, we omit the standard deviations in the testing results presented in the main text. In our experiments, we set $M$ equal to 8. Detailed testing results, including standard deviations, will be provided in the supplementary materials~\footnote{Supplementary materials are available at~https://github.com\\/ndwsfkba/EventMamba}.\\
\textbf{Hilbert Space-Filling Curve Mamba.}
In terms of temporal modeling, although some methods have attempted to facilitate channel-wise modeling by applying the same operations as in the spatial domain to the channel dimension of feature maps or using the reshape function in PyTorch to serialize event features, these methods inevitably lead to the loss of spatio-temporal correlation among adjacent pixels. To overcome this challenge, drawing inspiration from~\cite{wu2024rainmamba}, we introduce the Hilbert Space-Filling Curve Mamba (HSFCMamba), which utilizes space-filling curves to convert unstructured event features into regular sequences as shown in Figure~\ref{pic:2}.
Specifically, we select two representative space-filling curves, the Hilbert curve and its transposed variant Trans-Hilbert curve, to scan event features. Compared to ordinary reshape function scanning curves, space-filling curves like the Hilbert curve exhibits superior locality preserving properties and a lower space-to-linear ratio~\cite{chen2022efficient}, meaning that adjacent keypoints in the scanned one-dimensional sequence typically have geometrically close positions in the three-dimensional space. We believe this property can largely retain the spatial relationships among points, which is essential for accurate feature representation and analysis in event data. As a complement, the Trans-Hilbert curve exhibits similar characteristics but scans from a different perspective, providing a diversified view of spatial locality.
Concretely, we use the following expression for serialization:
\begin{equation}
\begin{split}
&z_h, \bar{z}_h = \text{MLP}(\text{HSFC}(\hat{E}_{v-1})+\text{pos\_embedding}), \\
&\hat{E}_l = \text{VMB}(\text{concat}(z_h, \bar{z}_h)),\\
\end{split}
\end{equation}
where HSFC represents the Hilbert scan and Trans-Hilbert scan methods, $z_h$ and $\bar{z}_h$ is the serialized output respectively. Subsequently, we concatenate the sequences $z_h$ and $\bar{z}_h$ obtained from the two different scanning methods, each with a shape of $(B, \frac{H}{8} \times \frac{W}{8} \times 8C)$, into a total sequence $\hat{z}_l$ with a shape of $(B, \frac{H}{8} \times \frac{W}{8} \times 8C \times 2)$. The sequence $\hat{z}_l$ is then fed into the VMB module and reshaped through Hilbert/Trans-Hilbert index to obtain its original shape, constructing the enhanced spatio-temporal event features $\hat{E}_l$. Through our HSFCMamba, our network better preserves spatio-temporal correlations among event data.

\begin{table*}[htbp] 
\small
  \centering
  \begin{tabularx}{\textwidth}{p{2.71cm}|p{1.2cm}<{\centering}p{1.2cm}<{\centering}p{1.2cm}<{\centering}|p{1.2cm}<{\centering}p{1.2cm}<{\centering}p{1.2cm}<{\centering}|p{1.2cm}<{\centering}p{1.2cm}<{\centering}p{1.2cm}<{\centering}}
  \hline
  \multirow{2}{*}{Method} & \multicolumn{3}{c|}{HQF} & \multicolumn{3}{c|}{IJRR} & \multicolumn{3}{c}{MVSEC} \\
  \cline{2-10}
  & MSE$\downarrow$ & SSIM$\uparrow$ & LPIPS$\downarrow$ & MSE$\downarrow$ & SSIM$\uparrow$ & LPIPS$\downarrow$ & MSE$\downarrow$ & SSIM$\uparrow$ & LPIPS$\downarrow$ \\
  \hline
  E2VID+ & 0.036 & 0.533 & \textBF{0.252} & 0.055 & 0.518 & \underline{0.261} & 0.132 & 0.264 & 0.514\\
  FireNet+ & 0.041 & 0.471 & 0.316 & 0.062 & 0.464 & 0.318 & 0.218 & 0.212 & 0.569\\
  SPADE-E2VID & 0.077 & 0.400 & 0.486 & 0.079 & 0.462 & 0.422 & 0.138 & 0.266 & 0.589\\
  EVSNN & 0.065 & 0.424 & 0.502 & 0.093 & 0.413 & 0.531 & 0.149 & 0.253 & 0.576\\
  ET-Net &0.035 & 0.558 & 0.274 & 0.053 & 0.552 & 0.296 & 0.115 & 0.315 & 0.493\\
  HyperE2VID &\underline{0.033} & \underline{0.563} & 0.272 & \underline{0.047} & \underline{0.569} & 0.278 & \underline{0.096} & \underline{0.319} & \underline{0.489}\\
  \hline
  EventMamba &\textBF{0.031} &\textBF{0.575} &\underline{0.261} &\textBF{0.039} & \textBF{0.586}& \textBF{0.254}& \textBF{0.073}&\textBF{0.328}&\textBF{0.475} \\
  \hline
\end{tabularx}
  \caption{Comparison on HQF, IJRR and MVSEC Datasets. Best and second best indexs are marked in bold and underline.}
\label{tab:1}
\end{table*}

\section{Experiments}
\subsection{Settings}
\textbf{Implementation Details.}
The network is trained for 400 epochs on an NVIDIA 3090 GPU using the AdamW optimizer~\cite{kingma2014adam} with a batch size of 4, a patch size of $128\times128$, an initial learning rate of 1e-4, and an exponential decay strategy with a gamma of 0.99.
\\
\textbf{Loss Functions.}
Our loss function over $t$-times can be written as:
\begin{equation}
\label{eq:21}
\mathcal{L}=\sum_{t=0}^{N_0} \mathcal{L}_{LPIPS}^t+\lambda_{TC} \sum_{t=L_0}^{N_0} \mathcal{L}_{temp}^t,
\end{equation}
where $\mathcal{L}_{LPIPS}$ represents the LPIPS loss\cite{zhang2018unreasonable}, and $\mathcal{L}_{temp}$ represents the temporal consistency loss~\cite{li2021enforcing,zhang2021learning}. A comprehensive description of these loss functions will be provided in the supplementary materials. The hyperparameters $N_0$, $L_0$, and $\lambda_{TC}$ are set to 20, 2, and 0.5, respectively.\\
\textbf{Training Datasets.}
We utilize the ESIM~\cite{rebecq2018esim} multi-object 2D renderer option to generate a synthetic training dataset. This renderer captured multiple moving objects within the 2D motion range of the camera. The dataset consists of 280 sequences, each with a duration of 10 seconds. The contrast threshold for event generation ranged from 0.1 to 1.5. Each sequence includes generated event streams, ground truth video frames, and optical flow maps, with an average frequency of 51 Hz. The resolution of both the event camera and frame camera is $256\times256$. These sequences encompass up to 30 foreground objects with varying velocities and trajectories, randomly selected from the MS-COCO dataset~\cite{lin2014microsoft}.\\
\textbf{Testing Datasets.}
We compare the video quality of event stream reconstruction on three publicly available training datasets: HQF~\cite{stoffregen2020reducing}, IJRR~\cite{mueggler2017event}, and MVSEC~\cite{zhu2018multivehicle}.\\
\textbf{Evaluation Metrics.}
In order to quantitatively evaluate the structural quality of video reconstruction, we follow the approach of E2VID and use the following commonly used metrics: Mean Squared Error (MSE), Structural Similarity Index (SSIM)~\cite{hore2010image}, and Learned Perceptual Image Patch Similarity (LPIPS)~\cite{zhang2018unreasonable}.\\
\textbf{Evaluation Methods.}
In our comparative analysis, we evaluate the performance of EventMamba against multiple cutting-edge techniques in the field, namely FireNet+~\cite{scheerlinck2020fast}, E2VID+~\cite{stoffregen2020reducing}, ET-Net~\cite{weng2021event}, SPADE-E2VID~\cite{cadena2021spade}, EVSNN~\cite{zhu2022event}, and HyperE2VID~\cite{ercan2024hypere2vid}.

\subsection{Quantitative Comparison}
Table~\ref{tab:1} showcases the quantitative comparison results between our proposed EventMamba network and previous methods. In terms of MSE and SSIM values, our Mamba network surpasses all existing event-based video reconstruction networks. Regarding the LPIPS metric, our proposed method also outperforms the majority of existing approaches. However, there is a slight decrease compared to the E2VID+ method in the HQF dataset.
Our EventMamba demonstrates superior performance in MSE, SSIM, and LPIPS metrics, achieving state-of-the-art results across multiple datasets. Additionally, our EventMamba model shows greater computational efficiency than Transformer based method ET-Net, with detailed findings available in the~\textbf{Ablation Studies and Analysis}.

\begin{table}[tbp]
\small

  \centering
  \begin{tabularx}{0.48\textwidth}{p{1cm}|p{0.7cm}<{\centering}p{0.8cm}<{\centering}p{0.8cm}<{\centering}p{0.7cm}<{\centering}p{0.7cm}<{\centering}p{0.7cm}<{\centering}}
  \hline
  Config. & (a) & (b) & (c) & (d) & (e) & Ours\\
  \hline
  RWO & $\times$ & \checkmark & \checkmark & \checkmark & $\times$& \checkmark\\
  HSFC & \checkmark & Hilbert & Trans & $\times$ & $\times$ & \checkmark\\
  \hline
  MSE $\downarrow$ & 0.048 & 0.041 & 0.041 & 0.045 & 0.049 & 0.039\\
  SSIM$\uparrow$ & 0.562 & 0.582 & 0.581 & 0.574 & 0.559 & 0.586\\   
  \hline
\end{tabularx}
\caption{Ablation study on network structures on IJRR.}
\label{tab:2}
\end{table}

\begin{figure*}[htbp]
\centering
\includegraphics[width=\linewidth]{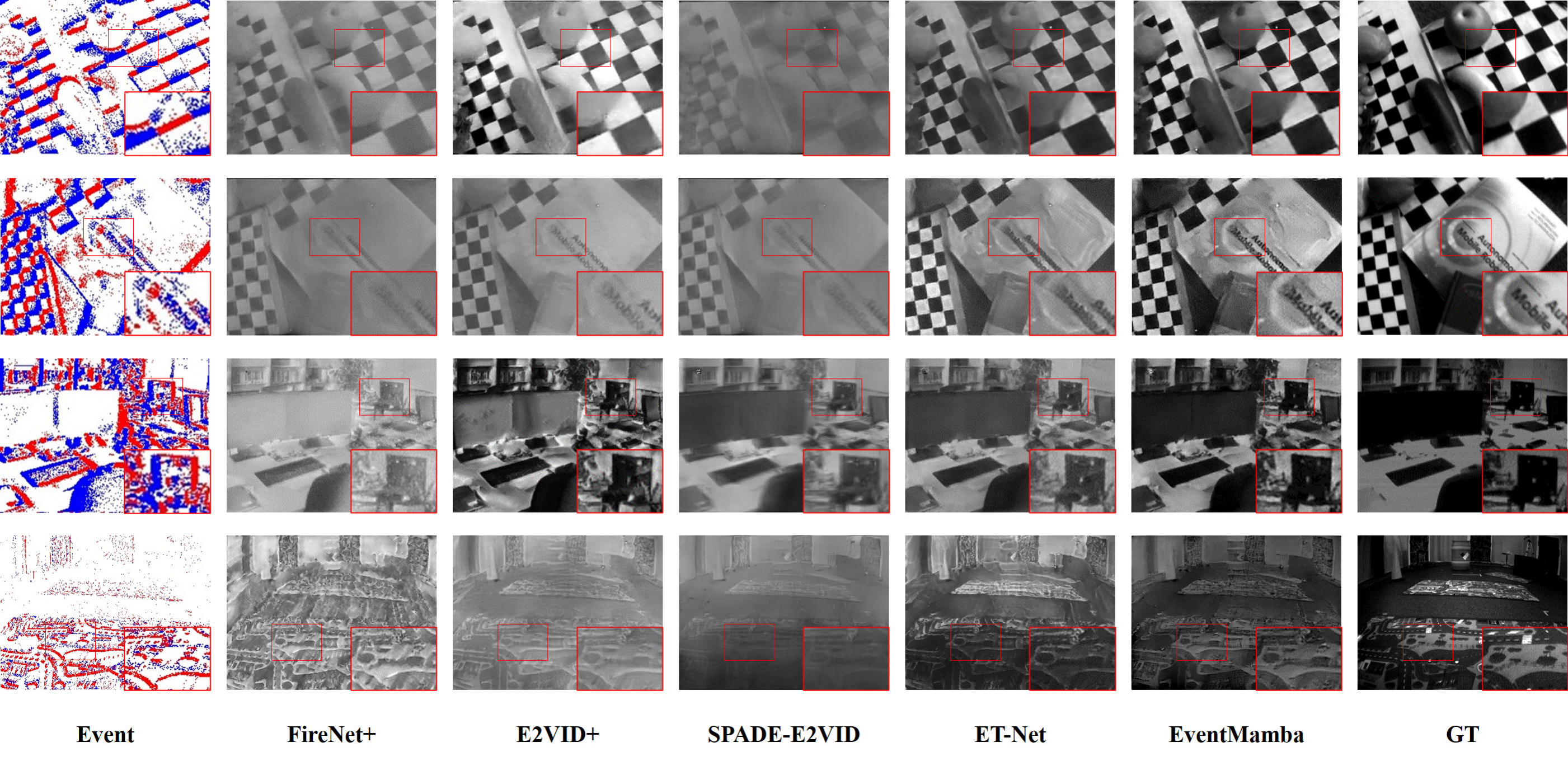}
\caption{Qualitative comparisons on three benchmarks from HQF (row1-2), IJRR (row3), and MVSEC (row4).}
\label{pic:3}
\end{figure*}

\begin{figure}
\centering
\includegraphics[width=\linewidth]{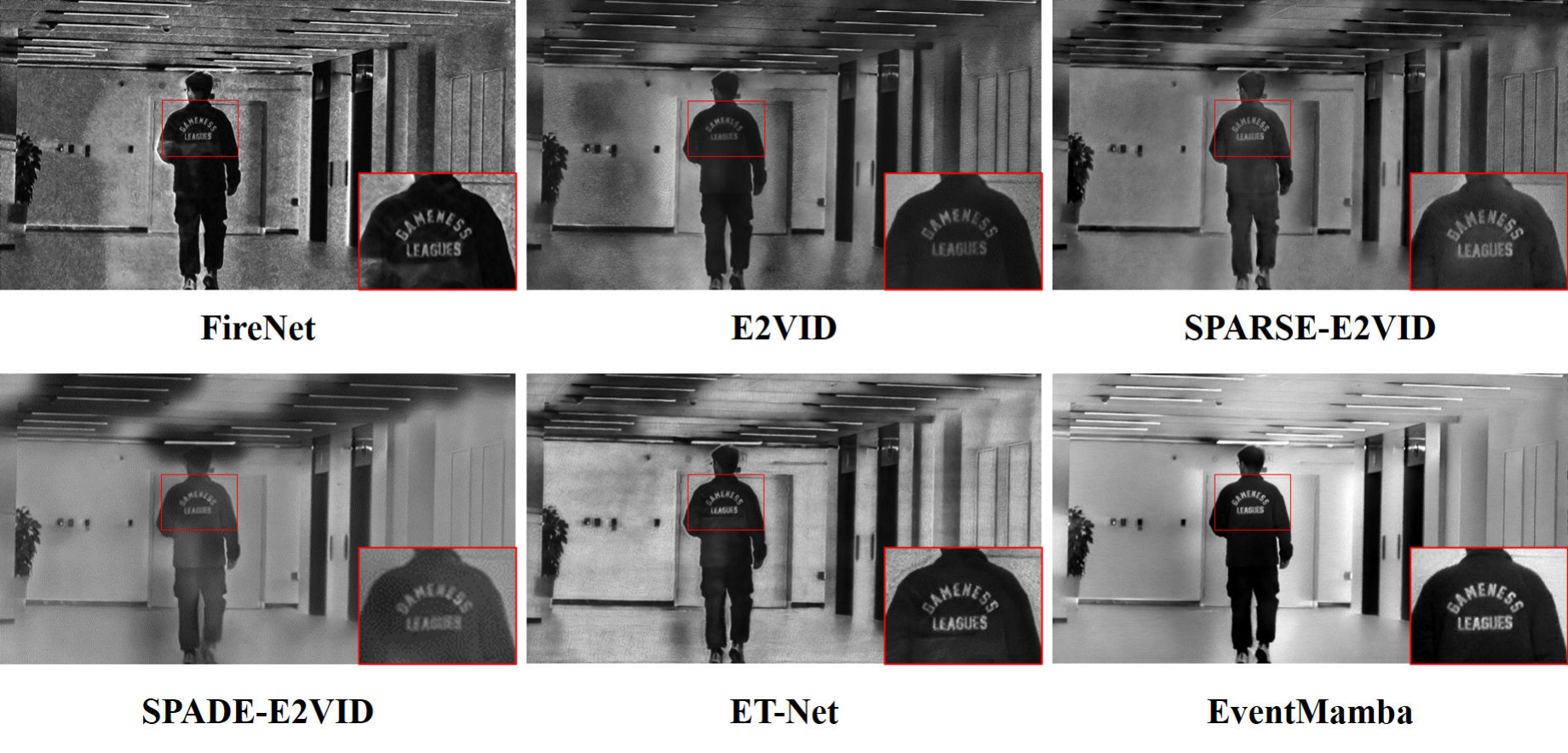}
\caption{Qualitative comparisons on sequences captured by the Prophesee EVK4 camera.}
\label{pic:4}
\end{figure}

\subsection{Qualitative Comparison}
Figure~\ref{pic:3} demonstrates the qualitative reconstruction results of our EventMamba and all baseline methods on images from video clips of the HQF, IJRR, and MVSEC datasets. Ground Truth (GT) video frames are also presented for comparison. It is observed that the frames reconstructed by FireNet+ and E2VID+ lack accuracy in brightness, leading to an inferior overall visual quality of the images. The reconstruction outcomes of ETNet and HyperE2VID are visually more appealing than those of FireNet+ and E2VID+, rendering a more lifelike scene. In contrast, our EventMamba further enriches the final reconstruction with more intricate details, whilst mitigating common artifacts seen in E2VID+. Furthermore, the image contrast of our reconstructed frames closely matches that of the GT images. These qualitative results corroborate the data presented in Table~\ref{tab:1}.
To further support our comparative analysis, we capture a series of high-definition video sequences using the Prophesee EVK4 camera. Figure~\ref{pic:4} showcases a series of comparison images. EventMamba delivers superior visual outcomes, exhibiting fewer artifacts than other methods.

\begin{table}[tbp] 
\small
  \centering
  \begin{tabularx}{0.47\textwidth}{p{2.3cm}|p{2.3cm}<{\centering}p{2.3cm}<{\centering}}
  \hline
  & Params (M) & Times (ms) \\
  \hline
  FireNet+ (L) & 0.04 & 11.8 \\
  FireNet+ (H) & 0.04 & 47.73 \\
  \hline
  E2VID+ (L) & 10.71 & 14.55 \\
  E2VID+ (H) & 10.71 & 81.82 \\
  \hline
  SPADE (L) & 11.46 & 24.09\\
  SPADE (H) & 11.46 & 210.45 \\
  \hline
  ET-Net (L) & 22.18 & 36.36\\
  ET-Net (H) & 22.18 & 1.85(s) \\
  \hline
  HyperE2VID (L) & 10.15 & 16.54\\
  HyperE2VID (H) & 10.15 & 126.93\\
  \hline
  Ours (L) & 11.21 & 18.82 \\
  Ours (H) & 11.21 & 136.14 \\
  \hline
\end{tabularx}
\caption{Computational complexity of different networks.}
\label{tab:3}
\end{table}

\subsection{Ablation Studies and Analysis}
Our ablation experiments are conducted on the IJRR dataset unless otherwise specified. More ablation studies and analysis are presented in the supplementary materials.\\
\textbf{Investigation of RWOMamba and HSFCMamba.}
To explore the impact of RWOMamba and HSFCMamba on experimental outcomes, we design several experimental configurations to validate the efficacy of the proposed methods, as shown in Table~\ref{tab:2}. We investigate different configurations including changing the window partition strategy in RWOMamba to fixed window partition (Config. (a)), using the Hilbert curve scan alone (Config. (b)), using the trans-Hilbert curve scan alone (Config. (c)), and the Pytorch reshape function scanning (Config. (d)). Config. (e) involves using fixed window partition along with the Pytorch reshape function scanning. The experimental results demonstrate that both RWOMamba and HSFCMamba contribute to significant performance improvements compared to the baseline configurations. These findings validate the effectiveness of our proposed methods in capturing and preserving spatio-temporal dependencies in event features, ultimately leading to enhanced outcomes in the EBVR task.\\
\textbf{Comparison of Computational Complexity.}\label{Comparison}
To verify the efficiency of our network, we analyze its computational complexity compared to other approaches, focusing on the total number of parameters and inference time on a 3090 GPU. To make our comparison more meaningful, we select resolutions of two common types of event camera sensors currently in use: low resolution $346 \times 240$ (L) and high resolution $1280 \times 720$ (H). Our experimental results are presented in Table~\ref{tab:3}. In the table, the model's parameter count is measured in millions (M), and inference time in milliseconds (ms). From Table~\ref{tab:3}, it is clear that our method strikes a good balance among computational complexity and performance. On one hand, compared to Transformer-based methods such as ET-Net, our method not only has half the parameters but also significantly reduces inference time when processing large-size inputs of $1280 \times 720$. On the other hand, compared to CNNs-based methods, our approach achieves substantial performance improvements with only a slight increase in computational costs, demonstrating the superiority of our proposed method.\\
\textbf{Investigation of Base Channel.}
We further analyze the effect of the base channel $C$ within our network  on the final outcomes. Quantitative comparisons with other methods are displayed in Table~\ref{tab:4}. The results demonstrate that setting the network's base number $C$ to 24 yields the highest performance compared to all prior methods. While increasing the base number beyond 32 offers some metric improvements, the gains begin to plateau. Consequently, we opt for a base channel of 32 to balance performance and efficiency.

\begin{table}[tbp]  
\small
  \centering
  \begin{tabularx}{0.46\textwidth}{p{2.2cm}|p{1.6cm}<{\centering}|p{1.3cm}<{\centering}p{1cm}<{\centering}}
  \hline
  Network & Params (M) & MSE$\downarrow$ & SSIM$\uparrow$\\
  \hline
  FireNet+ & 0.04 & 0.062 & 0.464\\
  E2VID+ & 10.71 & 0.055 & 0.518\\
  SPADE-E2VID & 11.46 & 0.079 & 0.462\\
  ET-Net & 22.18 & 0.053 & 0.553\\
  HyperE2VID & 10.15 & 0.047 & 0.569\\
  \hline
  Ours (8) & 0.70 & 0.063 & 0.493\\
  Ours (16) & 2.61 & 0.057 & 0.532\\
  Ours (24) & 5.86 & 0.047 & 0.568\\
  Ours (32) & 11.21 & \textBF{0.039} & \underline{0.586}\\
  Ours (48) & 22.06 & \textBF{0.039} & \textBF{0.587}\\
  \hline  
\end{tabularx}
   \caption{Ablation study on the number of base channels in terms of MSE/SSIM.}
\label{tab:4}
\end{table}

\section{Conclusion}
In this paper, we critically analyze the shortcomings of existing vision Mamba models in addressing Event-Based Video Reconstruction (EBVR) tasks. To better adapt vision Mamba models to the specific characteristics of EBVR tasks, we propose random window offset (RWO) and Hilbert space filling curve (HSFC) strategies in the spatial and temporal domains, respectively. Specifically, in the spatial domain, we replace fixed window partitioning during training with randomly offset window partitioning to ensure translation invariance. In the temporal domain, we employ Hilbert/trans-Hilbert scanning strategies for serialization to maintain the spatio-temporal locality of events. Based on these strategies, our proposed EventMamba model demonstrates outstanding performance on multiple datasets and achieves significant improvements in computational speed compared to Transformer-based models.

\section{Acknowledgments}
This work was supported by National Key R\&D Program of China under Grant 2020AAA0105702, National Natural Science Foundation of China (NSFC) under Grants 62225207, 62436008, 62422609 and 62276243.

\bibliography{aaai25}


\end{document}


\maketitle
In the supplementary materials, we will provide additional proofs for the methodology section, more ablation experiments, code implementations of core modules, and further visualization results.
\section{Advantages of Hilbert Curve}
We assert that the ordinary Hilbert curve proves to be more effective for data flattening than the PyTorch reshape function, as illustrated in Figure~\ref{pic:S1}. Our assertion is substantiated by examining how well the Hilbert curve preserves locality and maintains structural similarity between the flattened data and its original configuration.
\subsection{Advantage 1: Locality Preservation}
The locality of a space-filling curve is an important attribute and can be assessed by examining the segments it comprises~\cite{mokbel2003analysis}. Enhanced locality correlates with improved clustering of features.

\textbf{Definition 1 (Segments):}
\textit{In a space-filling curve, the segment connecting two consecutive points is referred to as a segment. For such a curve in a D-dimensional space with grid size $N$, there are $N^D-1$ segments linking $N^D$ elements.}

\textbf{Definition 2 (Jump Segments):}
\textit{In an N-dimensional space, for two consecutive points $P_i$ and $P_{i+1}$, if the distance between them exceeds 1 ($\lvert P_{i+1}-P_i \rvert > 1$), the segment connecting $P_i$ to $P_{i+1}$ is considered a Jump segment. This type of segment quantifies the number of jump connections.}

\textbf{Definition 3 (Still Segments):}
\textit{For two consecutive points $P_i$ and $P_{i+1}$ in an N-dimensional space, if the distance between them is zero in any dimension $k$ ($P_{i+1} = P_i$ in dimension $k$), the segment linking $P_i$ and $P_{i+1}$ is termed a Still segment in dimension $k$. A Still segment measures how much a space-filling curve needs to move in one dimension before it can advance in another.}

\begin{figure}
\centering
\includegraphics[width=\linewidth]{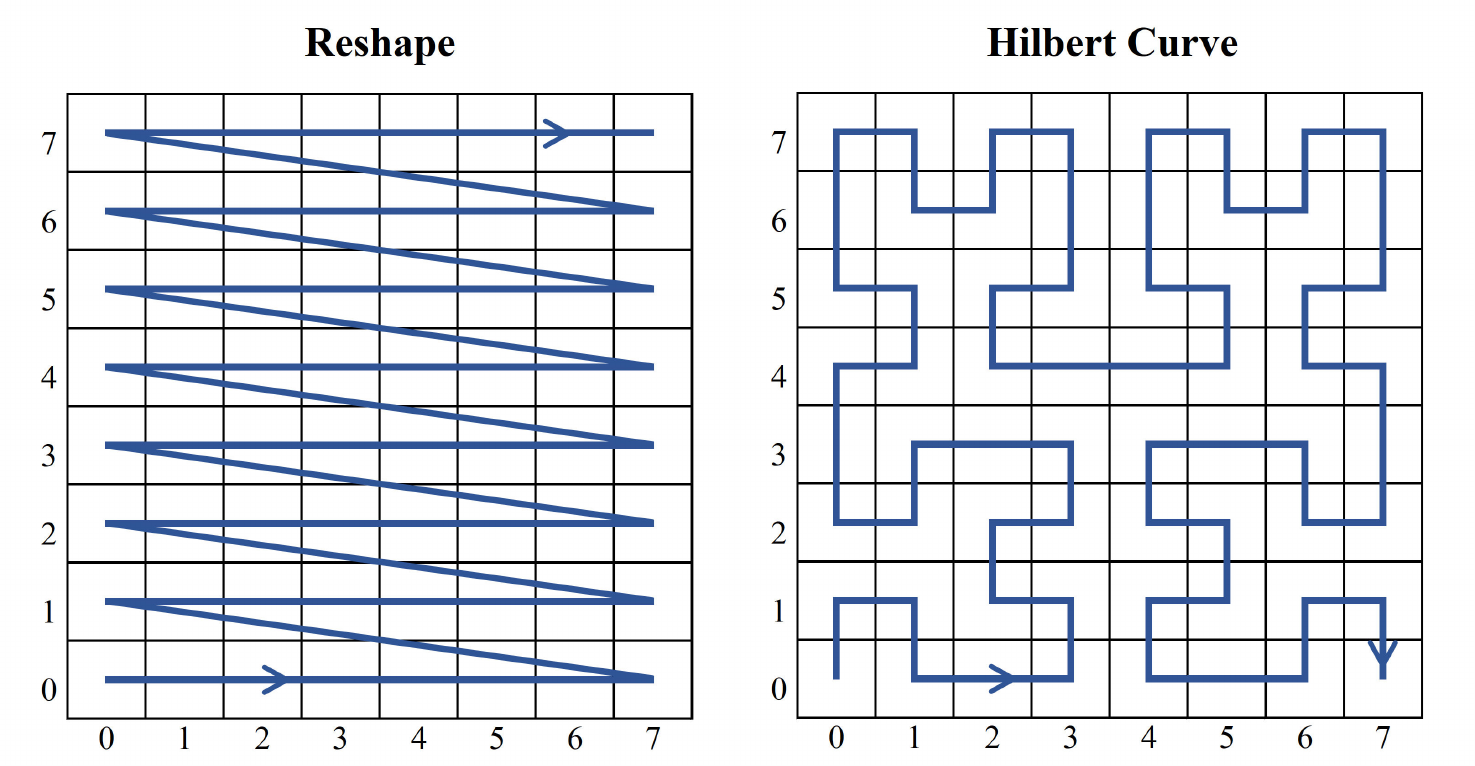}
\caption{Left: Mapping scheme of the Reshape function. Right: Mapping scheme of the Hilbert curve.}
\label{pic:S1}
\end{figure}

The Hilbert curve's most notable feature is its excellent ability to maintain locality, which significantly enhances the clustering of features. In Figure~\ref{pic:S1}, we visually compare its superiority over the reshape function. The subsequent theoretical justification involves assessing the locality of a space-filling curve through the proportion of \textit{Jump} and \textit{Still} segments. A lower proportion of \textit{Jump} segments indicates fewer discontinuous connections, while a higher proportion of \textit{Still} segments demonstrates greater consistency across dimensions. Both aspects contribute to improved locality, with calculations detailed in Equation~(\ref{eq:1}).
\begin{align}
\label{eq:1}
\begin{split}
&J_R=(\frac{N^D-1}{N-1}-D)\cdot\frac{1}{D(N^D-1)},\\
&S_R = (DN^D-N\frac{N^D-1}{N-1})\cdot\frac{1}{D(N^D-1)},\\
&J_H=0,\;S_H=(D-1)(N^D-1)\cdot\frac{1}{D(N^D-1)}.
\end{split}
\end{align}
Here, $J_R$ and $S_R$ denote the \textit{Jump} and \textit{Still} segments corresponding to the reshape function, while $J_H$ and $S_H$ correspond to those of the Hilbert curve. Additionally, $N$ and $D$ represent the grid size and dimensionality, respectively. The analysis shows that as $N$ increases, the proportion of \textit{Still} segments for the Hilbert curve remains consistently higher than that of the reshape function, and the \textit{Jump} segments for the Hilbert curve consistently register as zero. Consequently, the Hilbert curve is integrated into our framework to facilitate locality preserving .

\subsection{Advantage 2: Lower Space to Linear Ratio}
In addition to locality, the correspondence between the original structure and its flattened form is crucial because the flattening process inherently modifies the original configuration, resulting in previously contiguous points becoming separated once flattened. Consequently, the \textit{space to linear ratio} (SLR) is introduced to quantify the resemblance between the original structure and its linear representation.

\textbf{Definition 4 (Space to Linear Ratio):} \textit{When a pair of points, $p(t)$ and $p(\tau)$, within a two-dimensional space in $[0,1]\times[0,1]$ are mapped to corresponding one-dimensional points $t$ and $\tau$ in the interval $[0,1]$ using a space-filling curve $p$, defined by the function $p: [0,1] \to [0,1] \times [0,1]$, the ratio:}
\begin{equation}
\label{eq:2}
    \frac{|p(t)-p(\tau)|^2}{|t-\tau|},
\end{equation}
\textit{is termed the \textit{space to linear ratio} (SLR) of the two points.}

The maximum value reached by Equation (\ref{eq:2}) is referred to as the space to linear ratio (SLR) for the curve $p$. It is evident that a reduced SLR signifies a greater fidelity in the representation of the original structure by its flattened counterpart, thereby enhancing spatial locality.

\textbf{Theorem 1} \textit{The square-to-linear ratio of the Hilbert curve is equal to 6.}

The details of proof can be found in~\cite{bauman2006dilation}. While for some consecutive points in PyTorch reshape function, the SLR is $4^n-2^{n+1}+2$~\cite{zhao2022rethinking}, where $n$ is the curve order as defined before. It is obvious that Hilbert curve has lower SLR and therefore the flattened structure of Hilbert curve is closer to original one than PyTorch reshape function.

\begin{figure}
\centering
\includegraphics[width=\linewidth]{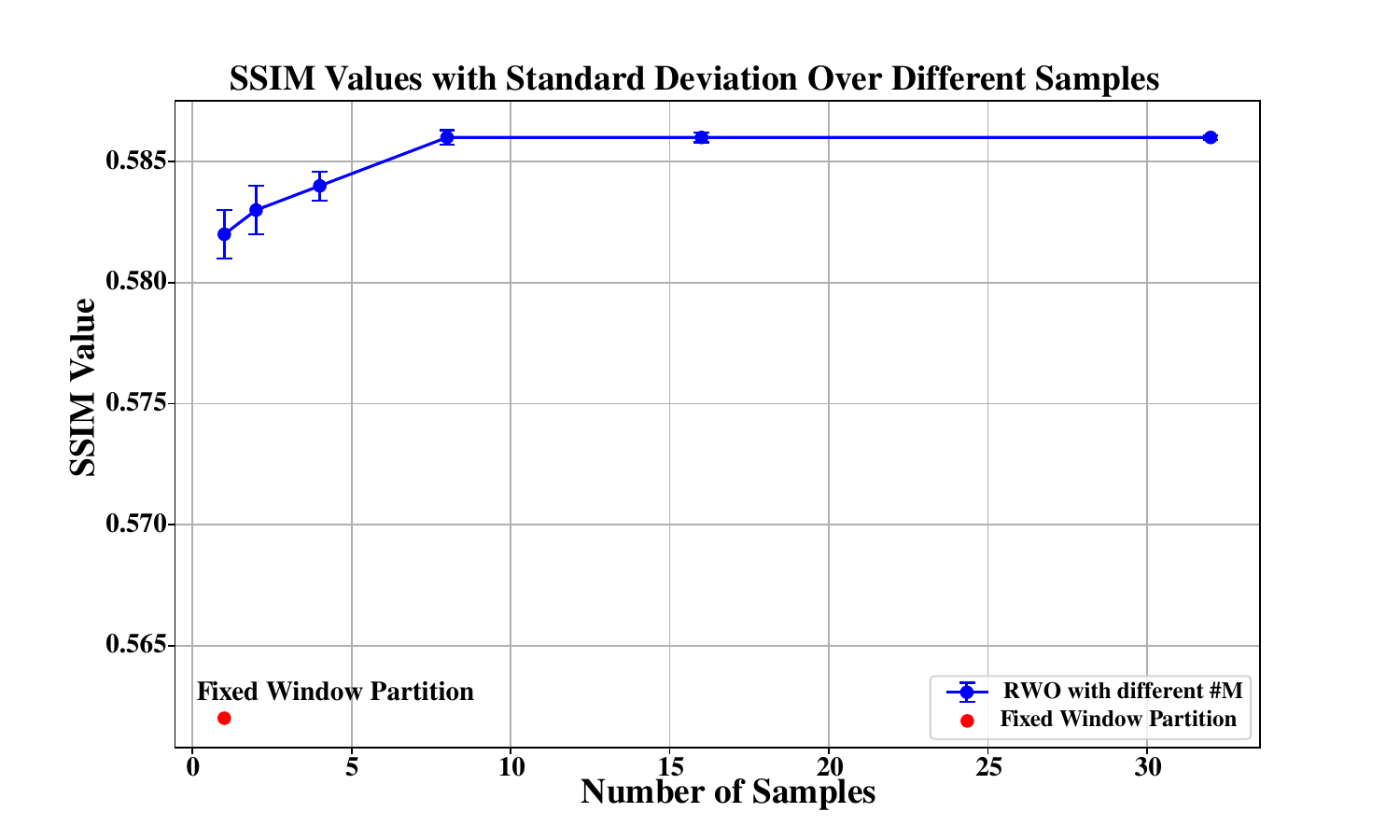}
\caption{SSIM vs. \# MC sample, the error bar stands for the standard deviation.}
\label{pic:S2}
\end{figure}

\begin{figure}
\centering
\includegraphics[width=\linewidth]{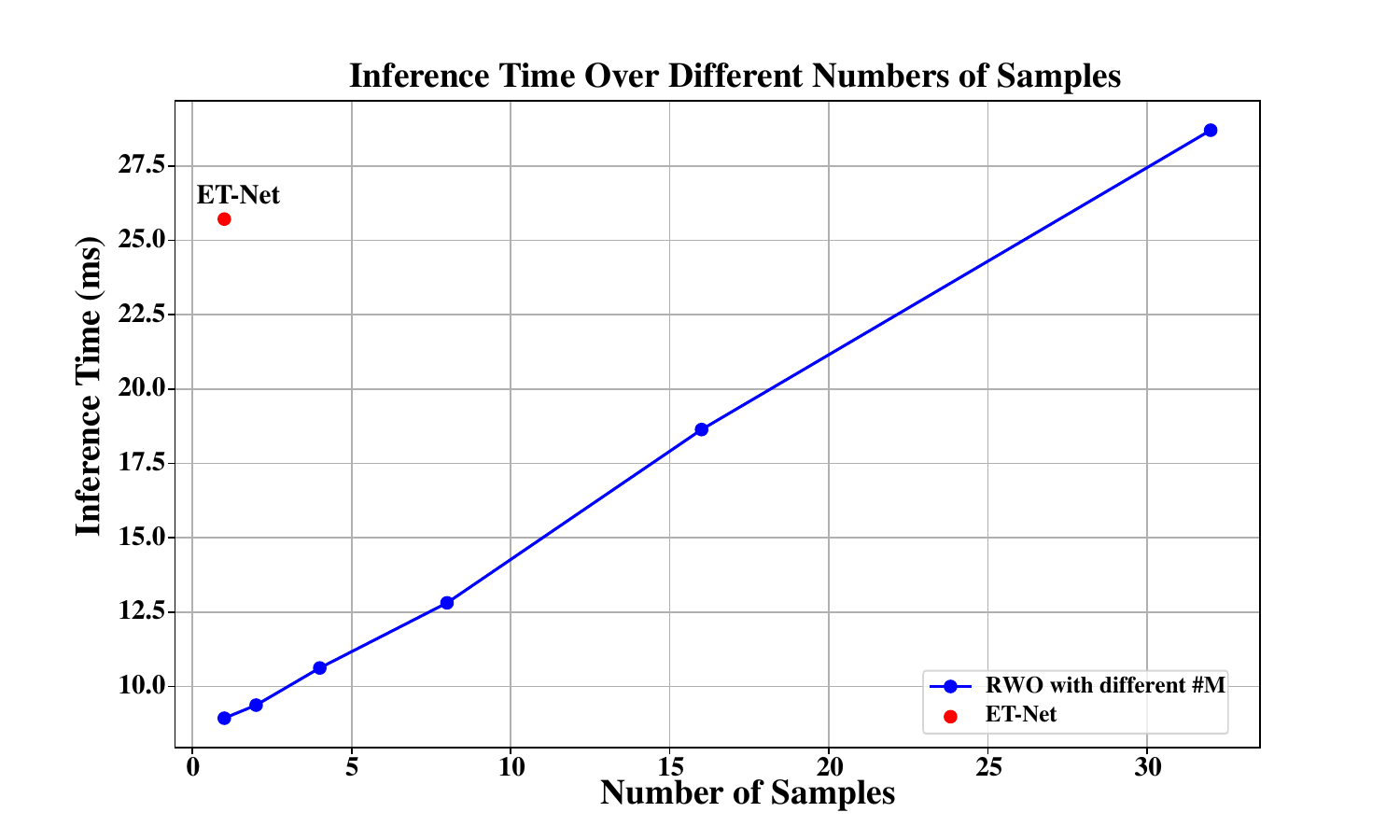}
\caption{Running time vs. \# MC sample.}
\label{pic:S3}
\end{figure}

\begin{table}[tbp]
  \caption{Ablation studies on loss functions.}
  \label{tab:2}
  \centering
  \begin{tabularx}{0.48\textwidth}{p{4.8cm}|p{1.2cm}<{\centering}p{1.2cm}<{\centering}}
  \hline
  Setting & MSE & SSIM\\
  \hline
  Without $\mathcal{L}_{temp}$ & 0.045 & 0.575\\
  $\mathcal{L}_{LPIPS}$ using AlexNet & 0.040 & 0.583\\
  $\mathcal{L}_{LPIPS}$ using VGGNet (original) & 0.039 & 0.586\\
  $\mathcal{L}_{temp}$ with $N=5$ & 0.042 & 0.581\\
  $\mathcal{L}_{temp}$ with $N=10$ & 0.041 & 0.583\\
  $\mathcal{L}_{temp}$ with $N=20$ & 0.039 & 0.586\\
  $\mathcal{L}_{temp}$ with $N=30$ & 0.039 & 0.586\\
  $\mathcal{L}_{temp}$ with $N=40$ & 0.039 & 0.585\\
  \hline
\end{tabularx}
\end{table}

\begin{table}[tbp]
  \caption{Ablation studies on different space-filling curves.}
  \label{tab:comparison2}
  \centering
  \begin{tabularx}{0.48\textwidth}{p{4.8cm}|p{1.2cm}<{\centering}p{1.2cm}<{\centering}}
  \hline
  SFC Setting & MSE & SSIM\\
  \hline
  Z-order Curve & 0.042 & 0.579\\
  Reshape Curve & 0.045 & 0.574\\
  Peano Curve & 0.042 & 0.580\\
  Hilbert Curve & 0.041 & 0.582\\
  Trans-Hilbert Curve & 0.041 & 0.581\\
  \hline
  Ours & 0.039 & 0.586\\
  \hline
\end{tabularx}
\end{table}

\begin{table}[tbp]
  \caption{Performance comparison at different event counts}
  \label{tab:event_sparsity}
  \centering
  \begin{tabularx}{0.48\textwidth}{X|>{\centering\arraybackslash}p{1.0cm}|>{\centering\arraybackslash}p{1.2cm}|>{\centering\arraybackslash}p{1.2cm}|>{\centering\arraybackslash}p{1.2cm}}
  \hline
  \textbf{Settings} & \textbf{5k} & \textbf{15k} & \textbf{25k} & \textbf{35k} \\
  \hline
  HyperE2VID      & 0.522       & 0.553        & 0.572        & 0.563        \\
  ETNet           & 0.519       & 0.548        & 0.560        & 0.559        \\
  Ours            & 0.536       & 0.563        & 0.581        & 0.578        \\
  \hline
\end{tabularx}
\end{table}

\begin{table}[tbp]
  \caption{Performance comparison at different levels of Gaussian noise}
  \label{tab:gaussian_noise}
  \centering
  \begin{tabularx}{0.5\textwidth}{X|>{\centering\arraybackslash}p{1.0cm}|>{\centering\arraybackslash}p{1.2cm}|>{\centering\arraybackslash}p{1.2cm}|>{\centering\arraybackslash}p{1.2cm}}
  \hline
  \textbf{Noise level $\sigma$ } & \textbf{$0.05$} & \textbf{$0.15$} & \textbf{$0.35$} & \textbf{$0.50$} \\
  \hline
  HyperE2VID      & 0.566                  & 0.464                  & 0.389                  & 0.316                  \\
  ETNet           & 0.548                  & 0.468                  & 0.396                  & 0.321                  \\
  Ours            & 0.584                  & 0.526                  & 0.442                  & 0.397                  \\
  \hline
\end{tabularx}
\end{table}

\section{More Ablation studies and Visualization}
\subsection{Determination of parameters for the Monte Carlo method}
To investigate the effectiveness of the Monte Carlo method in EBVR tasks, we design a series of experiments. Specifically, we test the impact of different numbers of Monte Carlo samples on model performance and resource consumption using the IJRR dataset. Each experimental setting is repeated 5 times to ensure the reliability of the results. Figure~\ref{pic:S2} and Figure~\ref{pic:S3} showcase the resulting curves obtained from the experiments. Through analysis, we draw two conclusions: First, as the number of samples increases, both the model's performance and resource consumption exhibit an upward trend. Balancing the value changes in the trade-off curve, we select $M=8$ samples as the equilibrium point between performance and efficiency. Second, even when using only a single sample, our method's efficiency is superior to traditional fixed-window partition methods, yet it achieves significant performance improvements. These findings validate the effectiveness and flexibility of the Monte Carlo method in our task.

\subsection{Investigation of loss functions}
To determine the final form of the loss function, we conduct ablation studies on the structure of the loss function and the selection of hyperparameters, as shown in Table~\ref{tab:2}. From the table, it is evident that the introduction of $\mathcal{L}_{temp}$ enhances the quality of reconstructed video frames. Additionally, we explore the network architecture within $\mathcal{L}_{LPIPS}$. It is apparent that VGGNet is more suitable than AlexNet as the network structure within $\mathcal{L}_{LPIPS}$. We also investigate the impact of $N_0$ on the experimental results, noting that there is no further improvement in the metrics when $N_0$ exceeds 20. Therefore, we select 20 as our experimental setting.

\subsection{Investigation of different space filling curves}
To verify the effectiveness of the proposed HSFCMamba, we conduct ablation experiments on various space-filling curves, including the Z-order curve~\cite{orenstein1986spatial}, PyTorch reshape curve, and the Peano curve~\cite{sagan1992geometrization}. The results of these experiments are presented in Table 2. The experimental structure demonstrates that the Hilbert curve is more suitable as a space-filling curve in the EBVR task. Our method, compared to the traditional single Hilbert scanning approach, offers improved spatial locality, thereby achieving the best experimental outcomes.

\subsection{Robustness test}
We evaluate the robustness of our method across various dimensions and compare its performance with previous state-of-the-art methods. Our tests focus on how well the method performs under different degrees of event sparsity and varying levels of Gaussian noise. The results, presented in Table~\ref{tab:event_sparsity} and Table~\ref{tab:gaussian_noise}, utilize the IJRR dataset and the SSIM index to measure performance. These findings clearly demonstrate that our method consistently outperforms existing technologies, such as the HyperE2VID method, in environments characterized by noise and varying levels of event tensor sparsity.

\subsection{Quantitative Results for Datasets Under Extreme Conditions}
To validate the video reconstruction capabilities of EventMamba under extreme conditions, we conduct experiments on the ECD\_fast and MVSEC\_night datasets. Since these two datasets lack high-quality ground truth (GT) references, we choose two no-reference metrics, the Blind/Referenceless Image Spatial Quality Evaluator (BRISQUE)~\cite{mittal2012no} and the Natural Image Quality Evaluator (NIQE)~\cite{mittal2012making}, as our evaluation criteria. Table~\ref{tab:extreme} shows that EventMamba generally outperforms other methods in these metrics, although it slightly trails in the BRISQUE score on the MVSEC\_night dataset.

\begin{table}[tbp]
  \caption{Ablation studies on extreme condition datasets.}
  \label{tab:extreme}
  \centering
  \begin{tabularx}{0.48\textwidth}{p{2cm}|p{2.4cm}<{\centering}p{2.4cm}<{\centering}}
  \hline
  Methods & ECD\_fast (BRISQUE/NIQE) & MVSEC\_night (BRISQUE/NIQE)\\
  \hline
  ET-Net & 19.70, 7.53 & 15.53, 5.23\\
  HyperE2VID & 14.02, 6.67 & 5.95, 5.09\\
  Ours & 13.74 , 6.31 & 6.02, 4.63\\
  \hline
\end{tabularx}
\end{table}

\subsection{Visualization Results}
\begin{figure*}
\centering
\includegraphics[width=\linewidth]{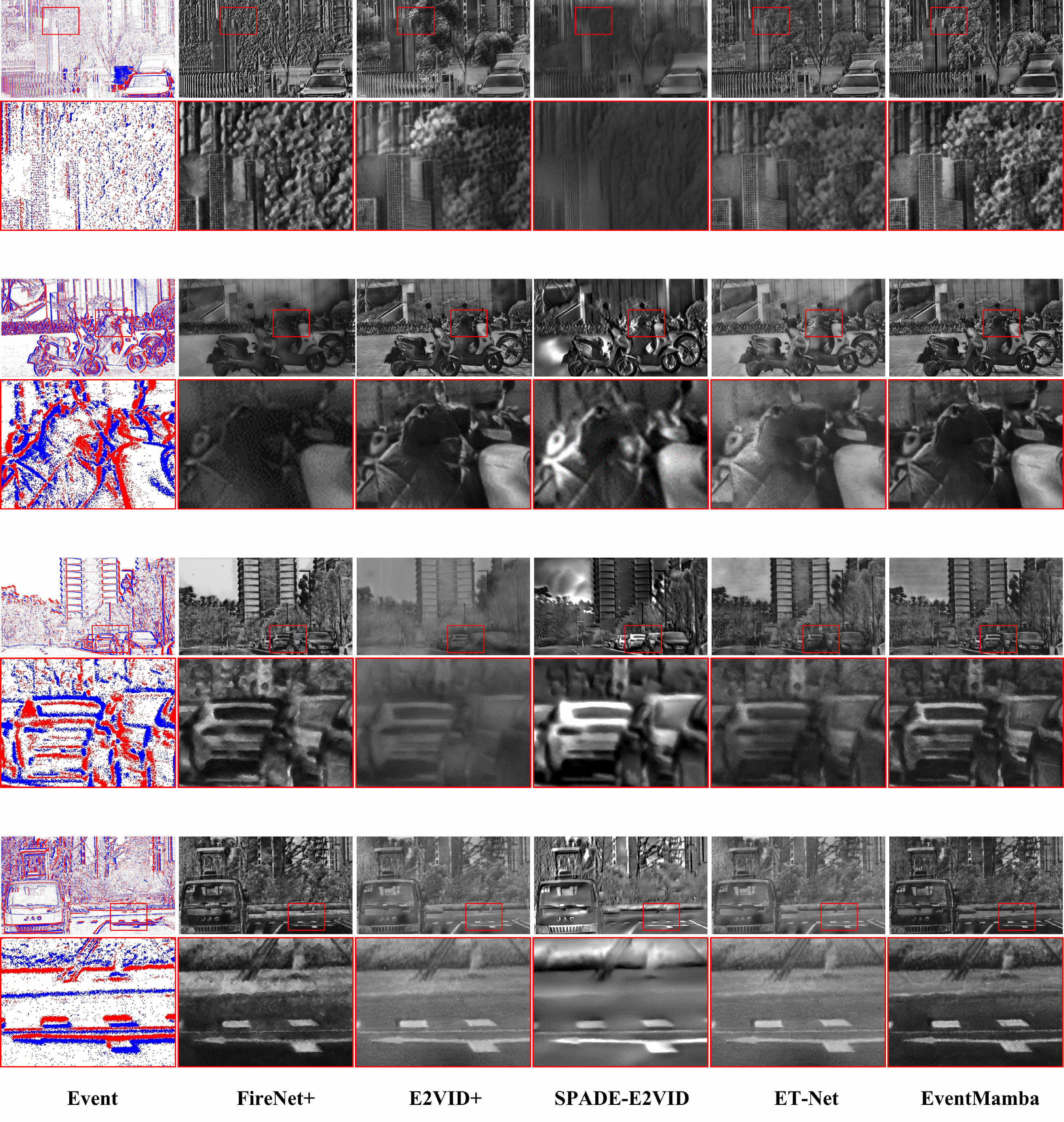}
\caption{Visualization comparisons on sequences captured by the Prophesee EVK4 camera.}
\label{pic:S4}
\end{figure*}

In this section, we visualize additional images captured by Prophesee EVK4 camera to demonstrate the reconstruction effectiveness of EventMamba. As shown in Figure~\ref{pic:S4}, experimental results indicate that our EventMamba achieves visual enhancements over previous methods.  In addition, we supplement our visualization results with a video to provide a more comprehensive visualization which can be found in the appendix.

\section{Implementation of Proposed methods}
For better understanding of the working mechanism of our proposed method, we also provide the code implementation of our proposed algorithm, which is presented in Algorithm~\ref{alg:1} and Algorithm~\ref{alg:2}.

\newpage
\bibliography{supp.bib}

\newpage
\begin{breakablealgorithm}
    \caption{Python Implementation of Hilbert Space Filling Curve}
    \label{alg:1}
    \begin{algorithmic}[1]
    \STATE \textcolor{red}{\textbf{def}} generate3d(x, y, z,
    \STATE \hspace{2.2cm}ax, ay, az, 
    \STATE \hspace{2.2cm}bx, by, bz,
    \STATE \hspace{2.2cm}cx, cy, cz):
    \STATE \hspace{0.6cm}w = abs(ax + ay + az)
    \STATE \hspace{0.6cm}h = abs(bx + by + bz)
    \STATE \hspace{0.6cm}d = abs(cx + cy + cz)
    \STATE \hspace{0.6cm}(dax, day, daz) = (sgn(ax), sgn(ay), sgn(az))\;\textcolor{cyan}{\textit{\# unit major direction ("right")}}
    \STATE \hspace{0.6cm}(dbx, dby, dbz) = (sgn(bx), sgn(by), sgn(bz))\;\textcolor{cyan}{\textit{\# unit ortho direction ("forward")}}
    \STATE \hspace{0.6cm}(dcx, dcy, dcz) = (sgn(cx), sgn(cy), sgn(cz))\;\textcolor{cyan}{\textit{\# unit ortho direction ("up")}}
    \STATE
    \STATE\hspace{0.6cm}\textcolor{cyan}{\textit{\# trivial row/column fills}}
    \STATE\hspace{0.6cm}if h == 1 and d == 1:
    \STATE\hspace{1.2cm}for i in range(0, w):
    \STATE\hspace{1.8cm}yield(x, y, z)
    \STATE\hspace{1.8cm}(x, y, z) = (x + dax, y + day, z + daz)
    \STATE\hspace{1.2cm}return
    
    \STATE\hspace{0.6cm}if w == 1 and d == 1:
    \STATE\hspace{1.2cm}for i in range(0, h):
    \STATE\hspace{1.8cm}yield(x, y, z)
    \STATE\hspace{1.8cm}(x, y, z) = (x + dbx, y + dby, z + dbz)
    \STATE\hspace{1.2cm}return
    
    \STATE\hspace{0.6cm}if w == 1 and h == 1:
    \STATE\hspace{1.2cm}for i in range(0, d):
    \STATE\hspace{1.8cm}yield(x, y, z)
    \STATE\hspace{1.8cm}(x, y, z) = (x + dcx, y + dcy, z + dcz)
    \STATE\hspace{1.2cm}return
    \STATE
    \STATE\hspace{0.6cm}(ax2, ay2, az2) = (ax//2, ay//2, az//2)
    \STATE\hspace{0.6cm}(bx2, by2, bz2) = (bx//2, by//2, bz//2)
    \STATE\hspace{0.6cm}(cx2, cy2, cz2) = (cx//2, cy//2, cz//2)
    \STATE
    \STATE\hspace{0.6cm}w2 = abs(ax2 + ay2 + az2)
    \STATE\hspace{0.6cm}h2 = abs(bx2 + by2 + bz2)
    \STATE\hspace{0.6cm}d2 = abs(cx2 + cy2 + cz2)
    \STATE
    \STATE\hspace{0.6cm}\textcolor{cyan}{\textit{\# prefer even steps}}
    \STATE\hspace{0.6cm}if (w2 \% 2) and (w $>$ 2):
    \STATE\hspace{1.2cm}(ax2, ay2, az2) = (ax2 + dax, ay2 + day, az2 + daz)
    \STATE\hspace{0.6cm}if (h2 \% 2) and (h $>$ 2):
    \STATE\hspace{1.2cm}(bx2, by2, bz2) = (bx2 + dbx, by2 + dby, bz2 + dbz)
    \STATE\hspace{0.6cm}if (d2 \% 2) and (d $>$ 2):
    \STATE\hspace{1.2cm}(cx2, cy2, cz2) = (cx2 + dcx, cy2 + dcy, cz2 + dcz)
    \STATE

    \STATE\hspace{0.6cm}\textcolor{cyan}{\textit{\# regular case, split in all w/h/d}}
    \STATE\hspace{0.6cm}else:
    \STATE\hspace{1.2cm}yield from generate3d(x, y, z,
    \STATE\hspace{4.4cm}bx2, by2, bz2,
    \STATE\hspace{4.4cm}cx2, cy2, cz2,
    \STATE\hspace{4.4cm}ax2, ay2, az2)
    
    \STATE\hspace{1.2cm}yield from generate3d(x+bx2, y+by2, z+bz2,
    \STATE\hspace{4.4cm}cx, cy, cz,
    \STATE\hspace{4.4cm}ax2, ay2, az2,
    \STATE\hspace{4.4cm}bx-bx2, by-by2, bz-bz2)
    
    \STATE\hspace{1.2cm}yield from generate3d(x+(bx2-dbx)+(cx-dcx),
    \STATE\hspace{4.4cm}y+(by2-dby)+(cy-dcy),
    \STATE\hspace{4.4cm}z+(bz2-dbz)+(cz-dcz),
    \STATE\hspace{4.4cm}ax, ay, az,
    \STATE\hspace{4.4cm}-bx2, -by2, -bz2,
    \STATE\hspace{4.4cm}-(cx-cx2), -(cy-cy2), -(cz-cz2))
    
    \STATE\hspace{1.2cm}yield from generate3d(x+(ax-dax)+bx2+(cx-dcx),
    \STATE\hspace{4.4cm}y+(ay-day)+by2+(cy-dcy),
    \STATE\hspace{4.4cm}z+(az-daz)+bz2+(cz-dcz),
    \STATE\hspace{4.4cm}-cx, -cy, -cz,
    \STATE\hspace{4.4cm}-(ax-ax2), -(ay-ay2), -(az-az2),
    \STATE\hspace{4.4cm}bx-bx2, by-by2, bz-bz2)
    
    \STATE\hspace{1.2cm}yield from generate3d(x+(ax-dax)+(bx2-dbx),
    \STATE\hspace{4.4cm}y+(ay-day)+(by2-dby),
    \STATE\hspace{4.4cm}z+(az-daz)+(bz2-dbz),
    \STATE\hspace{4.4cm}-bx2, -by2, -bz2,
    \STATE\hspace{4.4cm}cx2, cy2, cz2,
    \STATE\hspace{4.4cm}-(ax-ax2), -(ay-ay2), -(az-az2))
    \end{algorithmic}
\end{breakablealgorithm}

\begin{breakablealgorithm}
    \caption{Python Implementation of Random Window Offset Mask}
    \label{alg:2}
    \begin{algorithmic}[1]
    \STATE \textcolor{red}{\textbf{def}} random\_window\_offset\_mask(x, win\_size):
    \STATE \hspace{0.6cm}B, C, H, W = x.shape
    \STATE \hspace{0.6cm}H\_offset = random.randint(0, win\_size - 1)
    \STATE \hspace{0.6cm}W\_offset = random.randint(0, win\_size - 1)
    \STATE
    \STATE\hspace{0.6cm}shift\_mask = torch.zeros((B, 1, H, W)).type\_as(x)
    \STATE
    \STATE\hspace{0.6cm}if H\_offset $>$ 0:
    \STATE\hspace{1.2cm}h\_slices = (slice(0, -win\_size),
    \STATE\hspace{2.8cm}slice(-win\_size, -H\_offset),
    \STATE\hspace{2.8cm}slice(-H\_offset, None))
    \STATE\hspace{0.6cm}else:
    \STATE\hspace{1.2cm}h\_slices = (slice(0, None),)
    \STATE
    \STATE\hspace{0.6cm}if W\_offset $>$ 0:
    \STATE\hspace{1.2cm}w\_slices = (slice(0, -win\_size),
    \STATE\hspace{2.8cm}slice(-win\_size, -W\_offset),
    \STATE\hspace{2.8cm}slice(-W\_offset, None))
    \STATE\hspace{0.6cm}else:
    \STATE\hspace{1.2cm}w\_slices = (slice(0, None),)
    \STATE
    \STATE\hspace{0.6cm}cnt = 0
    \STATE\hspace{0.6cm}for h in h\_slices:
    \STATE\hspace{1.2cm}for w in w\_slices:
    \STATE\hspace{1.8cm}shift\_mask[:, :, h, w] = cnt
    \STATE\hspace{1.8cm}cnt += 1
    \STATE\hspace{0.6cm}return shift\_mask
    \end{algorithmic}
\end{breakablealgorithm}